%% file: emnlp2020.tex
\documentclass[11pt,a4paper]{article}
\usepackage[hyperref]{emnlp2020}
\usepackage{times}
\usepackage{latexsym}

\usepackage{microtype}
\usepackage{times}
\usepackage{latexsym}
\usepackage{graphicx}
\usepackage{amsmath}
\usepackage{booktabs}
\usepackage{multirow}
\usepackage{subfigure}
\usepackage{verbatim}
\usepackage{algorithm}
\usepackage{algorithmic}
\usepackage{makecell}
\usepackage{CJKutf8}
\usepackage{flushend}
\aclfinalcopy 
\def\aclpaperid{1035} 

\title{Improving Attention Mechanism with Query-Value Interaction}

\author{Chuhan Wu$^\dagger$~~~~Fangzhao Wu$^\ddagger$~~~~Tao Qi$^\dagger$~~~~\textbf{Yongfeng Huang}$^\dagger$\\
    $^\dagger$Department of Electronic Engineering \& BNRist, Tsinghua University, Beijing 100084, China  \\
     $^\ddagger$Microsoft Research Asia, Beijing 100080, China\\
  \tt{\{wuchuhan15,wufangzhao,taoqi.qt\}@gmail.com} \\ \tt{yfhuang@tsinghua.edu.cn}
  }

\begin{document}

\maketitle

\begin{abstract}
Attention mechanism has played critical roles in various state-of-the-art NLP models such as Transformer and BERT.
It can be formulated as a ternary function that maps the input queries, keys and values into an output by using a summation of values weighted by the attention weights derived from the interactions between queries and keys. 
Similar with query-key interactions, there is also inherent relatedness between queries and values, and incorporating query-value interactions has the potential to enhance the output by learning customized values according to the characteristics of queries.
However, the query-value interactions are ignored by existing attention methods, which may be not optimal.
In this paper, we propose to improve the existing attention mechanism by incorporating query-value interactions.
We propose a query-value interaction function which can learn query-aware attention values, and combine them with the original values and attention weights to form the final output.
Extensive experiments on four datasets for different tasks show that our approach can consistently improve the performance of many attention-based models by incorporating query-value interactions.

\end{abstract}

\input{data/introduction.tex}

\input{data/method.tex}

\input{data/experiment.tex}

\input{data/conclusion.tex}

\bibliography{acl2020}
\bibliographystyle{acl_natbib}


\end{document}

%% file: data/introduction.tex
\section{Introduction}

Attention mechanism is a widely used technique in the NLP field~\cite{yang2016han}. 
Since its application to neural machine translation~\cite{bahdanau2015neural}, it has achieved great success by playing an important role in many state-of-the-art NLP models such as Transformer~\cite{vaswani2017attention} and BERT~\cite{devlin2019bert} that empower various tasks such as machine translation~\cite{gehring2017convolutional} and reading comprehension~\cite{wang2017gated}.
Thus, the improvement on attention mechanism would be beneficial for various NLP applications~\cite{ShenZLJPZ18}.

\begin{figure}[t]
	\centering
	\subfigure[Standard attention.]{
		\includegraphics[width=0.225\textwidth]{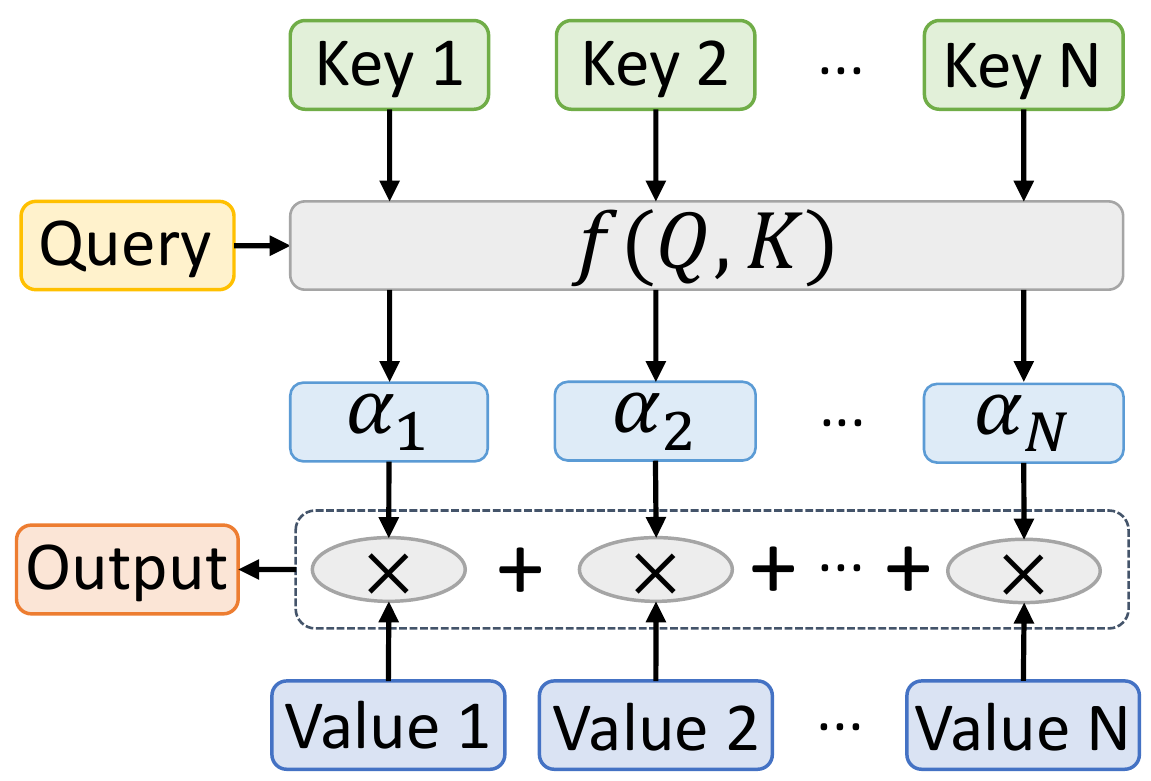}\label{fig.qvi}
		}
		\subfigure[Attention with QVI.]{
		\includegraphics[width=0.225\textwidth]{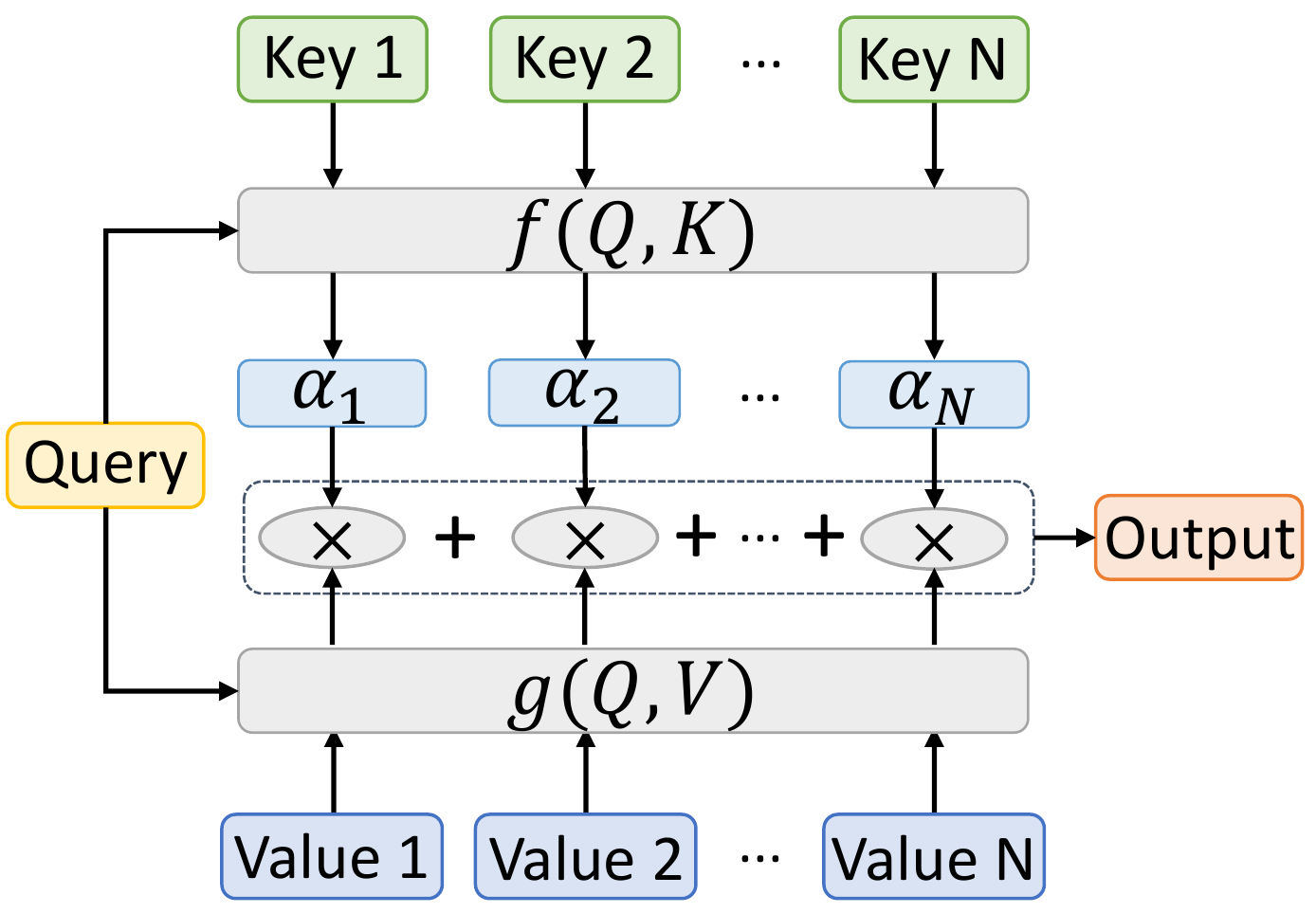}\label{fig.qvi2}
		}
	\caption{Standard attention and our proposed attention mechanism with query-value interactions (QVI).}\label{fig.exp}
\end{figure}

Attention mechanism is typically formulated as a ternary function, which maps the input queries ($\textbf{Q}$), keys ($\textbf{K}$) and values ($\textbf{V}$) to an output ($\textbf{O}$)~\cite{vaswani2017attention}.
In most attention networks the mapping has two steps, as shown in Fig.~\ref{fig.qvi}.
First, the attention network computes the attention weights  based on the interactions between the query and keys.
Then, the output is computed by the summation of values weighted by these attention weights.
To summarize, the output is formulated as $\textbf{O}=f(\textbf{Q}, \textbf{K})\textbf{V}$, where $f(\cdot)$ has many options such as dot product and perceptrons~\cite{vaswani2017attention}.
In this framework, the output is only derived from attention values and their attention weights computed by the interactions between queries and keys.
In fact, queries may have interactions with values, and incorporating the interactions between queries and values can adjust the values according to the characteristics of queries.
For example, in~\cite{wu2019npa} the attention network is used to learn representations of news for news recommendation, where the query is the embedding of a user and the values are the representations of words in the news clicked by this user. 
In this case, the query-value interactions can be used to adjust the word representations according to the user characteristics to learn better personalized news representations.
Thus, considering query-value interactions in attention mechanisms may be beneficial for learning more accurate output.
Motivated by these observations, in this paper we explore to extend attention mechanism from $\textbf{O}=f(\textbf{Q}, \textbf{K})\textbf{V}$ to $\textbf{O}=f(\textbf{Q}, \textbf{K})g(\textbf{Q},\textbf{V})$ by modeling the query-value interactions and explicitly encoding them into the output, as shown in Fig.~\ref{fig.qvi2}.

In this paper, we propose to improve existing attention mechanisms by modeling Query-Value Interactions (QVI), which can consider the relatedness between attention queries and values when forming the output.
More specifically, we propose a query-value interaction modeling function to learn query-aware attention values based on the element-level interactions between queries and values, and we further combine the query-aware attention values with the original values via gating mechanism.
The final output is formed by the combined values and the attention weights derived from the query-key interactions.
Extensive experiments on four benchmark datasets for two different tasks show that our approach can effectively improve the performance of various attention-based models by incorporating query-value interactions.

%% file: data/method.tex
\section{Preliminary on Attention Mechanism}
First, we give some brief introductions to attention mechanism. 
Additive attention and dot-product attention are two typical types of attention mechanisms~\cite{vaswani2017attention}.  
Their framework can be described as the function $\textbf{O}=f(\textbf{Q}, \textbf{K})\textbf{V}$. 
Additive attention usually aims to learn an output based on a summation of  input values, which takes their relative importance into account~\cite{yang2016han}.
In additive attention network, the query $\mathbf{q}$ is usually a parameter vector~\cite{wang2016attention} or an intermediate representation~\cite{ma2017interactive}, and the keys and values are usually a sequence with $N$ vectors, which is denoted as $\mathbf{v}=[\mathbf{v}_1, \mathbf{v}_2, ..., \mathbf{v}_N]$.
The additive attention network calculates an attention weight $\alpha_i$ for each value to indicate its relative importance using a function $f(\mathbf{q}, \mathbf{v}_i)$, which is formulated as follows:
\begin{equation}
    \alpha_i=f(\mathbf{q}, \mathbf{v}_i)=\frac{\exp(h(\mathbf{q},\mathbf{v}_i))}{\sum_{j=1}^{N}\exp(h(\mathbf{q},\mathbf{v}_j))},
\end{equation}
where $h(\cdot)$ is often implemented by dot product or perceptrons. 
The output $\mathbf{o}$ is a summation of input values weighted by their attention weights:
\begin{equation}
    \mathbf{o}=\sum_{i=1}^{N}\alpha_i\mathbf{v}_i=f(\mathbf{q}, \mathbf{v})\mathbf{v}.
    \label{att}
\end{equation}

Dot-product attention is another kind of attention mechanism that mainly relies on the product between the matrices of queries, keys and values~\cite{vaswani2017attention}.
In standard dot-product attention networks, the output  $\mathbf{O}$ is formulated as:
\begin{equation}
    \mathbf{O}=f(\mathbf{Q},\mathbf{K})\mathbf{V}=softmax(\frac{\mathbf{Q}\mathbf{K}^\top}{\sqrt{d}})\mathbf{V},
    \label{patt}
\end{equation}
where $\mathbf{Q}$, $\mathbf{K}$ and $\mathbf{V}$ are the query, key and value matrices, and $d$ is the dimension of their vectors.
In practice, one of the most widely used dot-product attention mechanism is self-attention, which is employed by various methods such as the Transformer~\cite{vaswani2017attention}, BERT~\cite{devlin2019bert} and XLNET~\cite{yang2019xlnet}.

\section{Attention Mechanism with Query-Value Interactions}\label{sec:Model}

In this section, we introduce the details of the proposed attention mechanism with Query-Value Interactions (QVI). 
Different from the standard attention mechanism with the function $\textbf{O}=f(\textbf{Q}, \textbf{K})\textbf{V}$, our proposed attention mechanism can be formulated as $\textbf{O}=f(\textbf{Q}, \textbf{K})g(\textbf{Q},\textbf{V})$.
Our method can be applied to both additive attention and dot-product attention mechanisms.
First, we introduce how to incorporate our QVI approach into additive attention.
In additive attention networks, the attention weights are computed by the the relevance between query and keys, and the output is a summation of values weighted by these attention weights. 
If the interactions between query and values can be modeled, we can adjust  values according to the characteristics of query, which may be useful for enhancing the output. 
To model the interactions between query and values, we propose to first apply a linear transformation to each value to align its dimension with the query, then we use the element-wise product between query and the transformed values to model their interactions at the element level to learn query-aware attention values.
Since values and their interactions with the query may have different relative importance for learning the output, we propose to use gating mechanism to aggregate query-aware values and the original values into a unified one.
The function $g(\mathbf{q},\mathbf{v}_i)$ that summarizes the process above is formulated as follows:
\begin{equation}
   g(\mathbf{q},\mathbf{v}_i)=(1-\beta_i)\mathbf{q}*\mathbf{W}\mathbf{v}_i+\beta_i \mathbf{v}_i,
  \label{fuseadd}
\end{equation}
\begin{equation}
   \beta_i=\sigma(\mathbf{u}^\top[\mathbf{q}*\mathbf{W}\mathbf{v}_i;\mathbf{v}_i]),
  \label{fuseadd1}
\end{equation}
where $*$ represents element-wise product, $\beta_i$ is an aggregating score that indicates the importance of the original value, $\mathbf{W}$ and $\mathbf{u}$ are parameters, and $\sigma(\cdot)$ is the sigmoid function.
The final output is computed as:
\begin{equation}
  \mathbf{\hat{o}}=\sum_{i=1}^{N}\alpha_i g(\mathbf{q},\mathbf{v}_i).
\end{equation}

Then, we introduce how to enhance dot-product attention with query-value interactions.
An intuitive way is to directly apply Eq.~(\ref{fuseadd}) to each pair of query and value vector.
However, it will harm the efficiency of dot-product attention since the computational cost is amplified by $N$.
Thus, we propose to compute a transformed query sequence $\mathbf{Q}$ as follows:
\begin{equation}
   \hat{\mathbf{Q}}=softmax(\frac{\mathbf{V}\mathbf{Q}^\top}{\sqrt{d}})\mathbf{Q},
    \label{dotlambda}
\end{equation}
where each vector in $\hat{\mathbf{Q}}$ is a weighted summation of all query vectors, and thereby we can use $\hat{\mathbf{Q}}$ and $\mathbf{V}$  to model query-value interactions without computing in a pair-wise manner. 
Similar to Eq.~(\ref{fuseadd}), the function $g(\mathbf{Q},\mathbf{V})$ that models the query-value interactions is formulated as follows:
\begin{equation}
   g(\mathbf{Q},\mathbf{V})=(1-\beta)\hat{\mathbf{Q}}*\mathbf{W}\mathbf{V}+\beta\mathbf{V},\label{fuseadd2}
\end{equation}
\begin{equation}
   \beta=\sigma(\mathbf{u}^\top[\hat{\mathbf{Q}}*\mathbf{W}\mathbf{V};\mathbf{V}]\mathbf{h})
\end{equation}
where $\beta$ is the aggregating score, * means element-wise product, and $\mathbf{W}$, $\mathbf{u}$ and $\mathbf{h}$ are parameters. 
The final output is computed as:
\begin{equation}
    \mathbf{\hat{O}}=softmax(\frac{\mathbf{Q}\mathbf{K}^\top}{\sqrt{d}})g(\mathbf{Q},\mathbf{V}).
    \label{datt}
\end{equation}
In our proposed attention mechanism, the attention values are customized according to the information of queries by incorporating query-value interactions, which can enhance the output.

%% file: data/experiment.tex
\section{Experiments}\label{sec:Experiments}

\subsection{Datasets and Experimental Settings}

We conducted experiments on two different tasks, i.e., text classification and named entity recognition (NER).
For text classification we use two datasets.
The first one is AG's News (denoted as \textit{AG})\footnote{https://www.di.unipi.it/en/}, which is a news topic classification dataset.
It contains 120,000 news articles for training and 7,600 news articles for test.
We randomly sampled 10\% of training samples as the validation set.
The second dataset is Amazon Electronics~\cite{he2016ups} (denoted as \textit{Amazon}), which is a widely used dataset for sentiment classification.
It contains 5 classes because the review ratings are ranged in [1, 5].
We randomly sample 40,000 reviews for training, 5,000 for validation and 5,000 for test.
For the NER task, we use two datasets provided by the SIGHAN Chinese language processing bakeoff (denoted as \textit{Bakeoff-3}\footnote{http://sighan.cs.uchicago.edu/bakeoff2006/download.html} and \textit{Bakeoff-4}\footnote{https://www.aclweb.org/mirror/ijcnlp08/sighan6/}), both of which are benchmark datasets for Chinese named entity recognition (NER).
The \textit{Bakeoff-3} dataset contains 46,364  sentences for training and 4,365 for test, and the \textit{Bakeoff-4} dataset contains 23,181 sentences for training and 4,636 for test.

In our experiments, the dimension of word embeddings was set to 300, and we used the pre-trained Glove embedding~\cite{pennington2014glove} to initialize the word embedding matrix.
In methods based on CNN, the number of filters was set to 256, and their window size was 3. 
In methods based on LSTM (bi-directional), the dimension of hidden representations was 256.
The complete settings of  hyperparameters are in supplements, and they were tuned according to the validation set.
For classification tasks, the metrics are accuracy and macro Fscore.
For the NER task, the metric is the micro Fscore.
We repeated each experiment 10 times and reported the average results.

\begin{table}[t]
	\centering
\resizebox{0.48\textwidth}{!}{
\begin{tabular}{|c|c|c|c|c|}
\hline
\multirow{2}{*}{\textbf{Methods}} & \multicolumn{2}{c|}{\textbf{AG}} & \multicolumn{2}{c|}{\textbf{Amazon}} \\ \cline{2-5} 
                                  & Accuracy        & Macro-F        & Accuracy          & Macro-F          \\ \hline
CNN                               & 92.18$\pm$0.11           & 92.16$\pm$0.11          & 64.88$\pm$0.43             & 42.94$\pm$0.41            \\
CNN-Att                           &  92.32$\pm$0.13               &   92.30$\pm$0.13             &   65.23$\pm$0.45                &    43.35$\pm$0.39              \\
CNN-Att-QVI                           &    92.66$\pm$0.10             &   92.65$\pm$0.10             &            65.89$\pm$0.40       &           44.04$\pm$0.38      \\ \hline
LSTM                     &    91.66$\pm$0.16               &     91.64$\pm$0.17           &       64.59$\pm$0.48            &      39.71$\pm$0.49            \\
LSTM-Att                &   92.20$\pm$0.14               &        92.18$\pm$0.14        &   67.05$\pm$0.42                &      43.74$\pm$0.40   \\
LSTM-Att-QVI                          &   92.68$\pm$0.12              &  92.65$\pm$0.12              &   67.42$\pm$0.44                &   44.34$\pm$0.41               \\ \hline
HAN                               &        92.12$\pm$0.09  &              92.10$\pm$0.10  &   67.19$\pm$0.39                &  44.98$\pm$0.41                \\
HAN-QVI                           &    92.74$\pm$0.10             &  92.71$\pm$0.11              &            \textbf{67.51}$\pm$0.40       &  \textbf{45.45}$\pm$0.37                \\ \hline
Transformer        & 93.11$\pm$0.12 & 93.09$\pm$0.13 & 65.15$\pm$0.42 & 42.14$\pm$0.40 \\
Transformer-QVI    & \textbf{93.40}$\pm$0.10 & \textbf{93.37}$\pm$0.10 & 65.82$\pm$0.35 & 43.20$\pm$0.37 \\
\hline
\end{tabular}
}
	\caption{The results on the \textit{AG} and \textit{Amazon} datasets.}\label{table.result}
\end{table}

\subsection{Experimental Results}

We verify the effectiveness of the proposed \textit{QVI} attention mechanism by comparing  many baseline methods and their variants with \textit{QVI} attention.
On the \textit{AG} and \textit{Amazon} datasets, the methods to be compared including:
(1) \textit{CNN}~\cite{kim2014convolutional}, convolutional neural networks;
(2)  \textit{CNN-Att}~\cite{gong2016hashtag}, applying additive attention after CNN;
(3)  \textit{CNN-Att-QVI}, using \textit{QVI} attention in \textit{CNN-Att};
(4)  \textit{LSTM}~\cite{hochreiter1997long}, long short-term memory network;
(5)  \textit{LSTM-Att}~\cite{zhou2016attention}, applying additive attention after LSTM;
(6)  \textit{LSTM-Att-QVI}, using \textit{QVI} attention in \textit{LSTM-Att};
(7)  \textit{HAN}~\cite{yang2016han}, a hierarchical attention network for document classification;
(8)  \textit{HAN-QVI}, using \textit{QVI} attention at both word and sentence levels;
(9)  \textit{Transformer}~\cite{vaswani2017attention}, using Transformer to learn text representations;
(10)  \textit{Transformer-QVI}, using \textit{QVI} attention in   Transformer.

The results on the \textit{AG} and \textit{Amazon} datasets are summarized in Table~\ref{table.result}.
From the results, we find that the methods with attention mechanisms perform better than their variants without attention.
It may be because attention mechanisms can model the informativeness and the interactions of words, which can help learn more accurate text representations.
In addition, compared with the methods using vanilla attention mechanisms (e.g., \textit{HAN} and \textit{Transformer}), their variants using \textit{QVI} attention (\textit{HAN-QVI} and \textit{Transformer-QVI}) perform better.
This may be because our \textit{QVI} approach can learn query-aware attention values according to the characteristics of attention query by modeling query-value interactions, which can enhance the output.

\begin{table}[t]
	\centering
\resizebox{0.48\textwidth}{!}{
\begin{tabular}{|c|l|l|c|l|l|c|}
\hline
\multirow{2}{*}{\textbf{Methods}} & \multicolumn{3}{c|}{\textbf{Bakeoff-3}}                                        & \multicolumn{3}{c|}{\textbf{Bakeoff-4}}                                        \\ \cline{2-7} 
                                  & \multicolumn{1}{c|}{\textbf{P}} & \multicolumn{1}{c|}{\textbf{R}} & \textbf{F} & \multicolumn{1}{c|}{\textbf{P}} & \multicolumn{1}{c|}{\textbf{R}} & \textbf{F} \\ \hline
Transformer-CRF                            & 85.77                           & 83.91                           & 84.83      & 86.35                           & 83.37                           & 84.82      \\
Transformer-QVI-CRF                         & 85.95                           & 84.67                           & 85.33      & 86.51                           & 84.09                           & 85.28      \\ \hline
CNN-Transformer-CRF                        & 86.77                           & 87.33                           & 87.04      & 86.56                           & 87.88                           & 87.24      \\
CNN-Transformer-QVI-CRF                     & 87.22                          & 87.64                           & \textbf{87.35}      & 86.95                           & 88.51                           & \textbf{87.70}      \\ \hline
\end{tabular}
}
	\caption{The performance of different methods on the \textit{Bakeoff-3} and \textit{Bakeoff-4} datasets. P, R, F represent precision, recall and Fscore, respectively. }\label{table.result2}
\end{table}

We also compare several methods on the SIGHAN \textit{Bakeoff-3} and \textit{Bakeoff-4} datasets, including:
(1) \textit{Transformer-CRF}~\cite{Cao000L18}, using Transformer to learn character representations and CRF to decode labels;
(2)  \textit{Transformer-QVI-CRF}, using \textit{QVI} mechanism in \textit{Transformer-CRF};
(3)  \textit{CNN-Transformer-CRF}, using a combination of CNN and Transformer;
(4)  \textit{CNN-Transformer-QVI-CRF}, using \textit{QVI} mechanism in  \textit{CNN-Transformer-CRF}.
The results on the \textit{Bakeoff-3} and \textit{Bakeoff-4} datasets are summarized in Table~\ref{table.result2}.
The results show that the proposed \textit{QVI} attention mechanism can consistently improve the performance of Transformer-based models in the NER task.

\begin{figure}[t]
	\centering
		\includegraphics[width=0.48\textwidth]{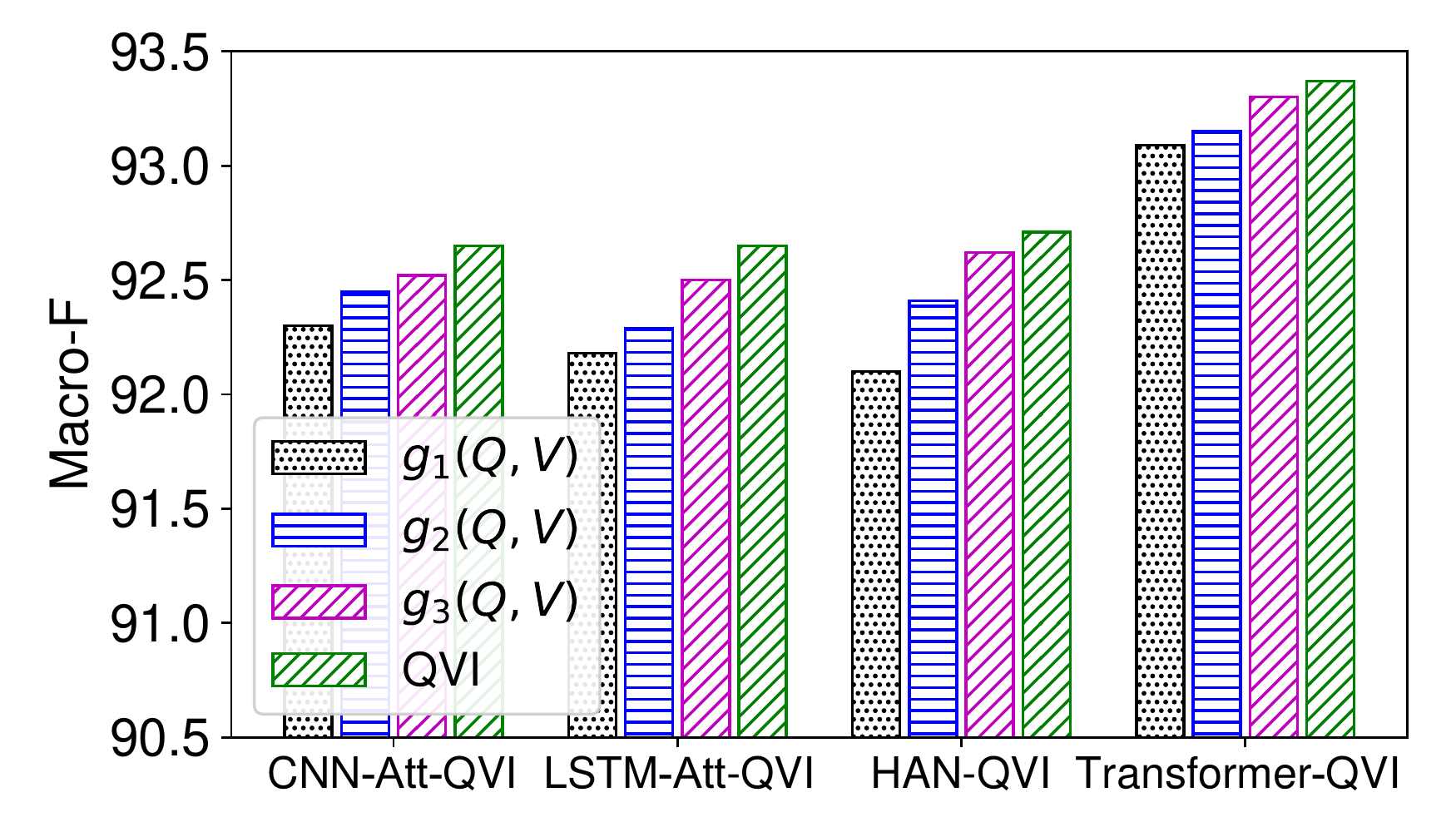}
\caption{Influence of QVI modeling.}\label{fig.fuse}
\end{figure}	

\subsection{Effect of Query-Value Interaction Modeling Methods}
We also study the influence of using different methods for modeling query-value interactions.
Concretely, we compare the proposed QVI attention with its variants with different $g(\mathbf{Q},\mathbf{V})$ functions, including: (1) values only: $g_1(\mathbf{Q},\mathbf{V})=\mathbf{V}$, (2) interactions only:  $g_2(\mathbf{Q},\mathbf{V})=\mathbf{Q}*\mathbf{W}\mathbf{V}$, (3) simple summation: $g_3(\mathbf{Q},\mathbf{V})=\mathbf{Q}*\mathbf{W}\mathbf{V}+\mathbf{V}$.
The performance of several attention-based methods using different $g(\mathbf{Q},\mathbf{V})$ on the \textit{AG} dataset is
shown in Fig.~\ref{fig.fuse}.
From the results, we find that the methods using query-value interactions are better than those using values only.
It shows that query-value interactions are very useful for learning more accurate output.
In addition, combining values with query-value interactions  is better than using interactions only.
This may be because the input values can be fully exploited in this way.
Besides, our QVI approach outperforms its variant without gating mechanism.
This may be because gating mechanism can model the relative importance of values and query-value interactions  to help learn more informative output.

%% file: data/conclusion.tex
\section{Conclusion}\label{sec:Conclusion}

In this paper, we propose to improve the existing attention mechanism by incorporating query-value interactions.
We propose a query-value interaction modeling function which can learn query-aware attention values, and further combine them with the original values and attention weights to derive the final output.
Extensive experiments on four benchmark datasets for text classification and NER show that the our approach can consistently improve the performance of  attention-based methods by modeling query-value interactions.
